\documentclass[lettersize,journal]{IEEEtran}

\usepackage{amsmath,amsfonts}
\usepackage{array}
\usepackage{textcomp}
\usepackage{stfloats}
\usepackage{url}
\usepackage{verbatim}
\usepackage{graphicx}
\usepackage{cite}
\usepackage[utf8]{inputenc}
\usepackage{epsfig}
\usepackage{graphicx}
\usepackage{subfigure}
\usepackage{tabularx}
\usepackage{amssymb}
\usepackage{amsmath}
\usepackage{amsthm} 
\usepackage{graphicx}
\usepackage{subcaption}
\usepackage{float}
\usepackage[ruled,vlined]{algorithm2e}
\usepackage{booktabs}
\usepackage{multirow}
\usepackage{color}
\usepackage{colortbl}
\usepackage{balance}
\usepackage{orcidlink}

\begin{document}

\title{PC-MNet: Dual-Level Congruity Modeling for Multimodal Sarcasm Detection via Polarity-Modulated Attention}

\author{
Maoheng Li\orcidlink{0000-0002-5331-2839},
Ling~Zhou\orcidlink{0000-0002-8313-5749},
Xiaohua~Huang\orcidlink{0000-0001-8897-3517},~\IEEEmembership{Senior~Member,~IEEE},
Rubing~Huang\orcidlink{0000-0002-1769-6126},~\IEEEmembership{Senior~Member,~IEEE},
Wenming~Zheng\orcidlink{0000-0002-7764-5179},~\IEEEmembership{Senior~Member,~IEEE},
and Guoying~Zhao\orcidlink{0000-0003-3694-206X},~\IEEEmembership{Fellow,~IEEE},
\thanks{Manuscript received April 19, 2026. This work was supported in part by the Science and Technology Development Fund of Macau, Macao SAR, under Grant Nos. 0021/2023/RIA1 and 0069/2025/RIB2. (\emph{Corresponding authors: Ling Zhou; Xiaohua Huang}.)}

\thanks{Maoheng Li and Ling Zhou are with the School of Computer Science and Engineering, Macau University of Science and Technology, Macao SAR 999078, China (e-mail: 3260006563@student.must.edu.mo; lzhou@must.edu.mo).}

\thanks{Xiaohua Huang is with the Oulu School, Nanjing Institute of Technology, Nanjing 210096, China (e-mail: xiaohuahwang@gmail.com).}

\thanks{Rubing Huang is with the School of Computer Science and Engineering, Macau University of Science and Technology, Macao SAR 999078, China, and also with the Macau University of Science and Technology Zhuhai MUST Science and Technology Research Institute, Zhuhai 519099, China (e-mail: rbhuang@must.edu.mo).}

\thanks{Wenming Zheng is with the Key Laboratory of Child Development and Learning Science (Southeast University), Ministry of Education, Nanjing 210096, China, and also with the School of Biological Science and Medical Engineering, Southeast University, Nanjing 210096, China (e-mail: wenming\_zheng@seu.edu.cn).}

\thanks{Guoying Zhao is with the Center for Machine Vision and Signal Analysis, University of Oulu, 90570 Oulu, Finland (e-mail: guoying.zhao@oulu.fi).}

}

\markboth{IEEE TRANSACTIONS ON MULTIMEDIA, 2026}
{How to Use the IEEEtran \LaTeX Templates}


\maketitle


\begin{abstract}
Multimodal sarcasm detection, which aims to precisely identify pragmatic incongruities between literal text and nonverbal cues, has gained substantial attention in multimodal understanding. Recent advancements have predominantly relied on na\"{\i}ve similarity-based attention mechanisms and uniform late fusion strategies. However, sarcastic utterances inherently exhibit polar-opposite semantic correlations rather than similarity. Consequently, employing similarity-seeking mechanisms inevitably fails to capture cross-modal contradictions, resulting in environmental noise, severe feature redundancy, and functional entanglement. To address these critical limitations, we propose the Polarity-Congruity Multimodal Network (PC-MNet), a novel hierarchical multi-granularity verification framework. Specifically, we first introduce a region-guided semantic alignment pipeline and a polarity-modulated attention mechanism to eradicate irrelevant visual noise and mathematically enforce the amplification of atomic-level cross-modal contradictions. This decoupling enables the extraction of the sracasm feature via parallel bipartite-dominant heterogeneous graphs. Furthermore, given that functional entanglement restricts traditional late fusions, we incorporate a scalar congruity routing mechanism and a prior-guided contextual graph. This mechanism anchors a generalized incongruity manifold through a two-stage asymmetric optimization driven by inconsistency-aware contrastive learning, selectively fusing only the most discriminative multi-granularity evidence. Extensive experiments on the \texttt{MUStARD} benchmark and its spurious-correlation-mitigated balanced datasets demonstrate that our approach achieves new state-of-the-art performance, surpassing the strongest multimodal baseline by a substantial 3.14\% improvement in Macro-F1. By architecturally isolating atomic, composition, and contextual conflicts. This work provides a robust, decoupled paradigm for modeling subtle pragmatic incongruities in human communication.
\end{abstract}



\section{Introduction}
\label{sec:introduction}

\begin{figure}[!b]
\centering
\includegraphics[width=0.485\textwidth]{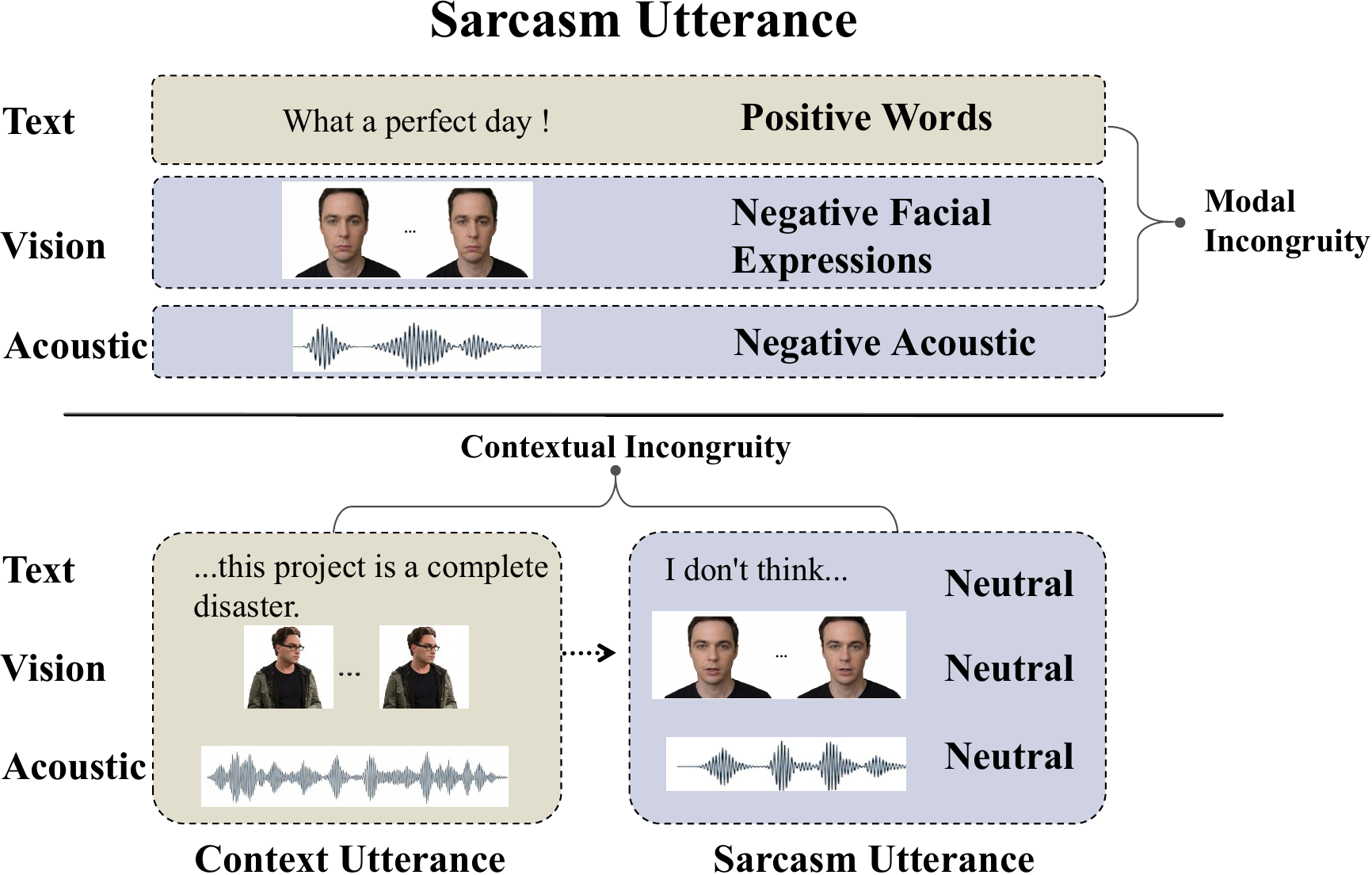}
\caption{Modal and contextual incongruity in multimodal sarcasm. The top example illustrates modal incongruity through conflicting polarities among text, visual, and acoustic signals within a single utterance. The bottom example demonstrates contextual incongruity, where a multimodally neutral utterance contradicts the preceding conversational context.}
\label{fig:motivation}
\end{figure}

\IEEEPARstart{M}{ultimodal} sentiment analysis plays a critical role in human-computer interaction and opinion mining. However, multimodal sarcasm detection remains highly challenging, due to the subtlety and ambiguity of pragmatic reasoning across different modalities. Sarcasm is a pragmatic phenomenon in which speakers convey sentiments contrary to their literal words \cite{campbell2012necessary}\cite{pan2024survey}. As shown in Figure \ref{fig:motivation}, prior studies \cite{ref3, ghosh2017contextual} identify two mechanisms of sarcasm: \textit{modal incongruity} and \textit{contextual incongruity}. Modal incongruity involves explicit cross-modal contradictions within an utterance, such as positive words accompanied by a dismissive tone \cite{ref38}. Conversely, contextual incongruity occurs when an internally consistent utterance contradicts the conversational context or speaker characteristics \cite{hasnat2022understanding}.

To address these challenges, current methods typically rely on feature fusion and interaction. Several studies employ tensor-based fusion \cite{ref7} or cross-modal attention \cite{ref38} to align multimodal sequences with text. Meanwhile, other approaches utilize Graph Neural Networks (GNNs) \cite{liang2022multimodal} to capture global conversational structures.

However, effectively modeling both forms of incongruity remains difficult due to two critical limitations. First, traditional cross-modal attention focuses on modality \textit{similarity} rather than \textit{contradiction}, often failing to capture the negative correlations that indicate sarcasm \cite{ref38}. Second, existing graph-based methods often entangle local intra-utterance conflicts with global inter-utterance dynamics. Consequently, this structural entanglement and the inherent lack of adaptability in dynamically shifting focus between modal and contextual incongruities severely limit the generalization capabilities of such models.

To alleviate these limitations, we propose the Polarity-Congruity Multimodal Network (PC-MNet), a hierarchical framework that explicitly captures contradictory multimodal cues. Unlike similarity-based methods, PC-MNet employs a decoupled reasoning pipeline to avoid feature homogenization. First, a region-guided semantic alignment module isolates human subjects to mitigate environmental noise and projects the refined visual cues into a cross-modally aligned semantic space to facilitate contradiction detection. Then, a polarity-modulated attention mechanism injects learnable polarity distances to capture fine-grained cross-modal contradictions. Furthermore, we construct parallel heterogeneous graphs to extract intra-utterance sarcasm features, offering an alternative to standard Transformer-based modeling. We also introduce a scalar congruity routing mechanism to isolate high-dimensional topological information, yielding scalar sarcasm priors. These priors then guide a downstream contextual graph to model conversational shifts. Finally, an adaptive dual-granularity fusion layer dynamically aggregates micro-level and macro-level evidence. The whole model is optimized via a two-stage strategy, transitioning from a valence-guided initialization to an inconsistency-aware contrastive loss.

In summary, our main contributions are as follows:
\begin{itemize}
    \item We propose PC-MNet, a hierarchical framework that addresses the feature entanglement problem in existing multimodal sarcasm detection models. By decoupling the reasoning pipeline, it effectively captures both micro-level modal and macro-level contextual incongruities.
    \item We propose a polarity-modulated attention mechanism and a prior-guided contextual graph neural network. These modules adaptively model contradiction-driven topologies, capturing both fine-grained cross-modal and inter-utterance contradictions.
    \item We design a scalar congruity routing mechanism to extract structural sarcasm priors. Additionally, we employ a two-stage optimization scheme with an inconsistency-aware contrastive loss to learn robust representations and prevent feature collapse.
    \item Experimental results on the \texttt{MUStARD} benchmark and the framework balanced datasets validate the effectiveness of PC-MNet. It outperforms existing fusion methods and large multimodal models, achieving a state-of-the-art Macro-averaged F1-score of 81.64\%, as supported by comprehensive ablation studies.
\end{itemize}

\section{Related Work}
\label{sec:related_work}

This section reviews the evolution of computational sarcasm detection from text-based methods to multimodal paradigms, highlighting the structural limitations that our approach aims at addressing.

\subsection{Multimodal Sentiment Analysis and Sarcasm Detection}
Early text-based sarcasm detection evolved from rule-based lexicons \cite{riloff2013sarcasm} to deep sequential models \cite{tay2018reasoning} and bidirectional transformers like BERT \cite{devlin2019bert}. However, text-only methods often fail when literal meanings are contradicted by nonverbal cues. Consequently, multimodal sentiment analysis emerged, utilizing early outer-product fusions \cite{ref7} or cross-attention and multi-view architectures \cite{ref38, yang2020image, wang2022cross} to optimize modality consistency. However, sarcasm inherently arises from affective incongruity \cite{campbell2012necessary}.

Regarding incongruity, extracting visual features directly from raw frames \cite{cai2019multi} or computing CLIP similarity can introduce environmental noise and semantic alignment bias. To address modality conflict, methods such as D\&R Net \cite{xu2020reasoning} and Att-BERT \cite{cai2019multi} employ discrepancy-aware attention. Nevertheless, relying on late-fusion strategies that rigidly concatenate intermediate representations often leads to feature redundancy. Furthermore, Multimodal Transformers that focus on similarity tend to homogenize features, obscuring essential cross-modal boundaries. Graph Neural Networks (GNNs) offer topological alternatives that preserve modality identities \cite{wu2025incongruity, liang2021cmgcn, ref_aaai24_graph}. Yet they often entangle modalities into monolithic graphs, leading to semantic over-smoothing and routing messages solely based on native feature similarity. Additionally, integrating external knowledge bases such as ConceptNet \cite{yue2023knowlenet} limits flexibility due to their static rules.

PC-MNet is designed to address these structural limitations. We use object detection to isolate human expressive features, thereby mitigating background noise. Instead of optimizing for alignment similarity, our polarity-modulated attention explicitly models cross-modal contradictions. To prevent similarity-based homogenization, we extract intra-utterance sarcasm features using parallel bipartite-dominant heterogeneous graphs, in which edges are weighted by polarity distances rather than simple feature similarity. Departing from late fusion, an adaptive dual-granularity fusion utilizes a scalar congruity routing mechanism to filter high-dimensional representations, using them exclusively as structural priors or contrastive optimization anchors. Finally, a two-stage asymmetric optimization uses continuous valence scores to warm up the model before applying contrastive supervision.

\subsection{Vision-Language Models and Fine-Grained Verification}
Recent benchmarks evaluate zero-shot and few-shot reasoning in large vision-language models using prompt-based paradigms for pragmatic deduction \cite{zhu2023prompt, liang2024fusion, zhang2025sarcasmbench, yue2026interarm}. Despite their scale, the autoregressive generation and flattened-sequence modeling inherent in these models limit their ability to perform explicit, fine-grained cross-modal verification. Without structural constraints to isolate micro-level polarity conflicts, these models are prone to hallucinations and precision-recall imbalances, often overpredicting sarcasm based solely on the global semantic context. PC-MNet addresses this architectural deficiency by introducing topological priors. By structurally decoupling micro-level conflicts, macro-level sarcasm features, and conversational dynamics, our network explicitly verifies fine-grained pragmatic incongruities rather than relying entirely on parameter scaling.

\begin{figure*}[htbp]
    \centering
    \includegraphics[width=0.9\textwidth]{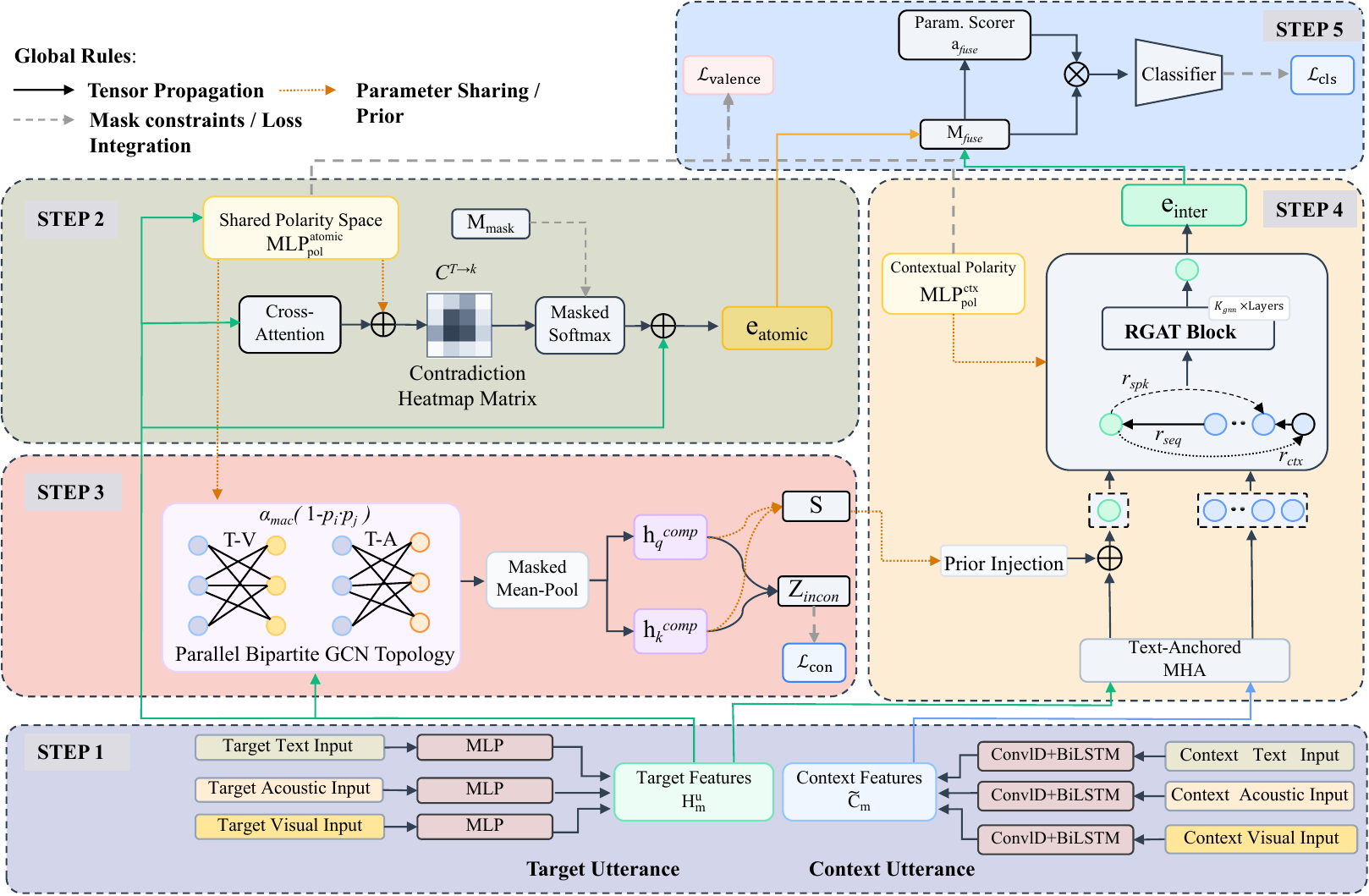}
    \caption{The overall architectural framework of PC-MNet.}
    \label{fig:framework}
\end{figure*}


\section{Methodology}
\label{sec:methodology}
The architecture of our Polarity-Congruity Multimodal Network is depicted in Figure \ref{fig:framework}, and the overall forward and optimization pipeline is formally summarized in Algorithm \ref{alg:pc-mnet_main}. Details of each constituent module are provided in the following sections.

\subsection{Problem Formulation}
\label{sec:task_description}
Formally, the framework receives a target utterance $\mathbf{U}_m$ across aligned modalities $m \in \{T, A, V\}$. Alongside the target utterance, the network processes the corresponding conversational history, denoted as $\mathbf{C}_m = [\mathbf{C}_{m,1}, \dots, \mathbf{C}_{m,J}]$, where $J$ represents the fixed context length. The primary objective is to predict the binary sarcasm label:
\begin{equation}\small 
\hat{y} = P(y = 1 \mid \{\mathbf{U}_m, \mathbf{C}_m\}_{m \in \{T,A,V\}}),
\end{equation}
where $y \in \{0, 1\}$ indicates a non-sarcastic or sarcastic state, respectively. Unlike conventional methods that blindly extract features, our approach utilizes explicit, continuous valence scores $v \in [-1, 1]$. Advanced multimodal benchmark annotations natively provide these valence scores. We utilize the native scores to enable a rigorous, context-aware cold-start for the polarity manifold.

\subsection{Step 1: Multimodal Feature Encoding}
\label{sec:step1}
We use BERT and Wav2Vec 2.0-base to extract sequential features $\mathbf{U}_T \in \mathbb{R}^{L_T \times d_T}$ and $\mathbf{U}_A \in \mathbb{R}^{L_A \times d_A}$. The variables $L_T$ and $L_A$ denote the sequence lengths of the textual tokens and acoustic frames, while $d_T$ and $d_A$ represent their respective original feature dimensions. To prevent background clutter from diluting pragmatic signals, a region-guided semantic alignment pipeline extracts focal subjects via the following two stages. Stage 1 uses YOLOv8 to detect human bounding boxes, thereby eliminating non-human noise and triggering deterministic zero-padding for human-absent frames. Stage 2 independently encodes retained crops via CLIP ViT-B/32, strictly prohibiting text-guided semantic filtering to preserve unadulterated crossmodal contradictions. The implemented detection-driven strategy effectively captures active speakers. Consequently, the pipeline excludes environmental noise while ensuring extracted visual embeddings $\mathbf{U}_V \in \mathbb{R}^{L_V \times K \times d_V}$ and context nodes $\mathbf{C}_{V,j} \in \mathbb{R}^{d_V}$, padded to maximum $K$ subjects per frame, remain authentic.

Applying a position-wise Multi-Layer Perceptron (MLP), we flatten visual spatio-temporal axes to $L_V^{eff} = L_V \times K$, mapping 2D sequences into a shared semantic space:
\begin{equation}\small
    \mathbf{H}_m = \text{Dropout}\left(\text{ReLU}\left(\text{LayerNorm}(\mathbf{U}_m \mathbf{W}_m + \mathbf{b}_m)\right)\right),
\label{eq:temporal_utterance}
\end{equation}
where $\mathbf{H}_m \in \mathbb{R}^{L_m^{eff} \times d_{enc}}$ represents the aligned sequence. For the historical context, we retain native sequences $\mathbf{H}_{m,j}$ for subsequent attention mechanisms while extracting utterance-level anchors via temporal mean-pooling: $\mathbf{h}_{m,j}^c = \text{MeanPool}(\mathbf{H}_{m,j}) \in \mathbb{R}^{d_{enc}}$.

\begin{algorithm}[t]
\caption{PC-MNet Forward \& Optimization Pipeline}
\small
\label{alg:pc-mnet_main}
\KwIn{Batch $\mathcal{B} = \{(\mathbf{U}_{m,i}, \mathbf{C}_{m,i}, y_i)\}_{i=1}^{|\mathcal{B}|}$ for $m \in \{T,A,V\}$, Stage parameter}
\KwOut{Predictions $\mathcal{Y}$, Total Loss $\mathcal{L}_{total}$}

$\mathcal{Y} \leftarrow \emptyset$\;
\For{each instance $i \in \mathcal{B}$}{
    Encode native sequences $\{\mathbf{H}_{m,i}\}$, history $\{\mathbf{H}_{m,j,i}\}$, and anchors $\{\mathbf{h}_{m,j,i}^c\}_{j=1}^J$\;

    $(\mathbf{e}_{atomic,i}, \mathbf{s}_{comp,i}, \mathbf{z}_{incon}^{(i)}) \leftarrow$ Atomic \& Composition Congruity described in Alg. \ref{alg:atomic_composition}\;
    
    Extract target base state $\mathbf{h}_{tgt,i}^{(0)}$ via temporal pooling\;

    $\mathbf{e}_{inter,i} \leftarrow$ Prior-Guided Contextual RGAT described in Alg. \ref{alg:inter_gnn}\;
    
    $\mathbf{M}_{fuse} \leftarrow [\text{Proj}(\mathbf{e}_{atomic,i}) ; \text{Proj}(\mathbf{e}_{inter,i})]$\;
    $\mathbf{a}_{fuse} \leftarrow \text{Softmax}(\mathbf{w}_{fuse}^\top \tanh(\mathbf{M}_{fuse} \mathbf{W}_F))$\;
    $\hat{y}_i \leftarrow \text{Classifier}(\mathbf{M}_{fuse}^\top \mathbf{a}_{fuse})$\;
    $\mathcal{Y} \leftarrow \mathcal{Y} \cup \{\hat{y}_i\}$\;
}
Compute $\mathcal{L}_{cls}$ using $\mathcal{Y}$ and ground truth $\{y_i\}$\;
Compute $\mathcal{L}_{con}$ using manifold representations $\{\mathbf{z}_{incon}^{(i)}\}$\;
\If{Stage == 1 (Cold-Start)}{
    Compute $\mathcal{L}_{valence}$ using native valence scores\;
    $\mathcal{L}_{total} \leftarrow \lambda_{cls}\mathcal{L}_{cls} + \lambda_{con}\mathcal{L}_{con} + \lambda_{val}\mathcal{L}_{valence}$\;
} \Else{
    $\mathcal{L}_{total} \leftarrow \lambda_{cls}\mathcal{L}_{cls} + \lambda_{con}\mathcal{L}_{con}$ 
}
\Return $\mathcal{Y}, \mathcal{L}_{total}$\;
\end{algorithm}

\subsection{Step 2: Atomic-Level Congruity}
\label{sec:atomic_level}
The detailed in Algorithm \ref{alg:atomic_composition}, explicitly modeling cross-modal contradictions begins by designating text as query $q=T$ and non-verbal modalities as keys $k \in \{A, V\}$, projecting sequences into a shared polarity space:
\begin{equation}\small
\begin{aligned}
    \mathbf{P}_T &= \text{L2Norm}(\text{MLP}_{pol}^{atomic}(\mathbf{H}_T)) \\
    \mathbf{P}_k &= \text{L2Norm}(\text{MLP}_{pol}^{atomic}(\mathbf{H}_k)).
\end{aligned}
\end{equation}
Sharing the weights of the $\text{MLP}_{pol}$ module across modalities ensures consistent angular measurements. During the initial cold-start phase, these projected polarities are directly supervised by continuous valence scores (as detailed in Section \ref{sec:fusion_and_loss}). Consequently, this explicit supervision guarantees that the formulated contradiction matrix $\mathbf{C}^{T \to k} = \mathbf{1} - \mathbf{P}_T \mathbf{P}_k^\top$ accurately quantifies the pragmatic divergence between verbal and non-verbal signals. Higher scores structurally modulate the attention score $\mathbf{S}$:
\begin{equation}\small
    \mathbf{S} = \text{MHA\_Scores}(\mathbf{H}_T, \mathbf{H}_k) + \alpha_{mic} \mathbf{C}^{T \to k}.
\end{equation}
Learnable amplifier $\alpha_{mic}$ controls micro-conflict sensitivity. Applying topological mask $\mathbf{M}_{mask}$ before Softmax and incorporating a residual connection yields the atomic-level conflict representation $\mathbf{E}_{atomic}^{T \to k} = \text{Softmax}(\mathbf{S} + \mathbf{M}_{mask}) \mathbf{H}_k + \mathbf{H}_T$, which routes attention to contradictory pairs while preserving intrinsic textual semantics. To extract the global micro-conflict signal, we apply temporal mean-pooling to both non-verbal branches and concatenate them, yielding the final unified atomic vector $\mathbf{e}_{atomic}$:
\begin{equation}\small
    \mathbf{e}_{atomic} = [\text{MeanPool}(\mathbf{E}_{atomic}^{T \to A}) \parallel \text{MeanPool}.(\mathbf{E}_{atomic}^{T \to V})].
\end{equation}

\begin{algorithm}[!t]
\caption{Atomic \& Composition Congruity}
\small
\label{alg:atomic_composition}
\KwIn{Target features $\mathbf{H}_T, \mathbf{H}_A, \mathbf{H}_V$}
\KwOut{Atomic vector $\mathbf{e}_{atomic}$, Tension prior $\mathbf{s}_{comp}$, Incongruity rep. $\mathbf{z}_{incon}$}

Project sequences to shared polarity space $\mathbf{P}_m, \forall m \in \{T,A,V\}$\;

\For{$k \in \{A, V\}$}{
    Calculate contradiction matrix $\mathbf{C}^{T \to k} \leftarrow \mathbf{1} - \mathbf{P}_T \mathbf{P}_k^\top$\;
    $\mathbf{E}_{atomic}^{T \to k} \leftarrow \text{PolarityModulatedMHA}(\mathbf{H}_T, \mathbf{H}_k, \mathbf{C}^{T \to k}, \alpha_{mic})$\;
}
$\mathbf{e}_{atomic} \leftarrow \text{MeanPool}(\mathbf{E}_{atomic}^{T \to A}) \parallel \text{MeanPool}(\mathbf{E}_{atomic}^{T \to V})$\;

\For{target pair $(q, k) \in \{(T, A), (T, V)\}$}{
    Initialize joint node matrix $\mathbf{H}_{joint}^{(0)}$ and polarity-modulated adjacency $\mathbf{A}_{mod}$\;
    $\mathbf{H}_{joint}^{(L_{mac})} \leftarrow \text{L-layer GCN}(\mathbf{H}_{joint}^{(0)}, \mathbf{A}_{mod})$\;
    Extract composition vectors $\mathbf{h}_q^{comp}, \mathbf{h}_k^{comp}$ from $\mathbf{H}_{joint}^{(L_{mac})}$\;
    Compute scalar tension $s_{comp}^{qk}$ and discrepancy vector $\boldsymbol{\Delta}_{qk}$\;
}
$\mathbf{s}_{comp} \leftarrow [s_{comp}^{TA}, s_{comp}^{TV}]$\;
$\mathbf{z}_{incon} \leftarrow \text{L2Norm}(\text{MLP}_{diff}([\boldsymbol{\Delta}_{TA} \parallel \boldsymbol{\Delta}_{TV}]))$\;

\Return $\mathbf{e}_{atomic}, \mathbf{s}_{comp}, \mathbf{z}_{incon}$\;
\end{algorithm}

\subsection{Step 3: Composition-Level Congruity}
\label{sec:composition_level}
To prevent the entanglement and over-smoothing inherent in dense Transformers, constructing parallel, bipartite-dominant heterogeneous graphs preserves identities and redefines message passing in response to polarity contradictions. Building these graphs exclusively for the target utterance focuses reasoning, prevents computational explosion, and defers historical context to Step 4 (Section \ref{sec:inter_sentence}), enforcing a scalar congruity routing. Injecting learnable modality-specific type embeddings $\mathbf{E}_{type}^q$ and $\mathbf{E}_{type}^k$ maintains identities:
\begin{equation}\small 
    \mathbf{H}_{joint}^{(0)} = [\mathbf{H}_q + \mathbf{E}_{type}^q ; \mathbf{H}_k + \mathbf{E}_{type}^k].
\end{equation}
Reusing the atomic polarity space ($\text{MLP}_{pol}^{atomic}$) established in Section \ref{sec:atomic_level}, projecting nodes into the shared manifold formulates the padding-masked adjacency matrix $\mathbf{A}_{mod}$ via a bounded amplifier $\alpha_{mac}$. The mechanism structurally forces the GCN to route messages across highly contradictory cross-modal nodes, actively amplifying heterophilous pragmatic conflicts without smoothing intrinsic spatial-temporal topologies. Assigning zero weights for padded interactions isolates invalid nodes, preventing semantic pollution during graph convolution.

After $L_{mac}$ layers, partitioning and masked mean pooling yield macro representations $\mathbf{h}_q^{comp}$ and $\mathbf{h}_k^{comp}$. Avoiding gradient interference from naive late fusion, a rigorous scalar congruity routing restricts composition-level features to two pathways. First, cosine congruity extracts purified macro-level features:
\begin{equation}\small 
    s_{comp}^{qk} = \text{CosineSim}(\mathbf{h}_q^{comp}, \mathbf{h}_k^{comp}).
\end{equation}
Routing the $O(1)$ prior $s_{comp}$ averts $O(d_{enc})$ fusion explosions. Meanwhile, branch features bypass final fusion for contrastive learning exclusively, and local sliding windows preserve intra-modal continuity.

\subsection{Step 4: Prior-Guided Contextual GNN}
\label{sec:inter_sentence}
Unlike standard conversational GNNs that blindly gather context, we introduce asymmetric node initialization and the injection of priors, as outlined in Algorithm \ref{alg:inter_gnn}. To circumvent background noise in historical utterances $j \in \{1, \dots, J\}$, text-anchored alignment designates pooled text $\mathbf{h}_{T,j}^c$ as a stable query extracting relevant cues from native acoustic ($\mathbf{H}_{A,j}$) and visual ($\mathbf{H}_{V,j}$) sequences via multi-head attention:
\begin{equation}\small
\begin{aligned}
    \mathbf{H}_{AV,j} &= [\mathbf{H}_{A,j} \parallel \mathbf{H}_{V,j}] \\
    \mathbf{h}_j^{(0)} &= \text{TextAnchored\_MHA}(\mathbf{Q}{=}\mathbf{h}_{T,j}^c, \mathbf{K}{=}\mathbf{H}_{AV,j}, \mathbf{V}{=}\mathbf{H}_{AV,j}).
\end{aligned}
\end{equation}
For the target utterance, distilling branch-specific features enforces the scalar congruity routing. We pool native sequences and project them to form the target base state $\mathbf{h}_{tgt}^{(0)}$. Injecting scalar prior $\mathbf{s}_{comp}$ via LayerNorm and $\mathbf{W}_{pri}$ creates the prior-aware target node $\tilde{\mathbf{h}}_{tgt}^{(0)} = \mathbf{h}_{tgt}^{(0)} + \text{LayerNorm}(\mathbf{W}_{pri} \mathbf{s}_{comp})$. Concatenating this prior-aware target state with the historical nodes constructs the initial conversational feature matrix $\mathbf{H}^{(0)} = [\mathbf{h}_1^{(0)}; \dots; \mathbf{h}_J^{(0)}; \tilde{\mathbf{h}}_{tgt}^{(0)}]$.

Three relations ($r_{seq}$, $r_{ctx}$, $r_{spk}$) yield binary masks $\mathbf{A}^r$. Projecting nodes into polarity space $\mathbf{p}_n$ computes opposition distance $C_{ij} = 1 - \mathbf{p}_i \cdot \mathbf{p}_j$, forming relational score $S_{ij,r} = \text{LeakyReLU}(\text{Linear}(\mathbf{H}_i, \mathbf{H}_j) + \alpha_{ctx} C_{ij})$. Normalizing over local neighborhood $\mathcal{N}_i^r$ via masked softmax isolates valid pairs for relation weights $\alpha_{ij,r} = \text{MaskedSoftmax}(S_{ij,r} \times \mathbf{A}_{i,j}^r)$. Message passing updates features: $\mathbf{H}_i^{(k)} = \text{ReLU}(\sum_{r} \sum_{j} \alpha_{ij,r} \mathbf{W}_r \mathbf{H}_j^{(k-1)})$. After $K_{gnn}$ layers, the target node forms contextual embedding $\mathbf{e}_{inter} = \mathbf{H}_{J+1}^{(K_{gnn})}$.

\begin{algorithm}[!t]
\caption{Prior-Guided Contextual RGAT}
\small
\label{alg:inter_gnn}
\KwIn{History anchors and sequences $\{\mathbf{h}_{T,j}^c, \mathbf{H}_{A,j}, \mathbf{H}_{V,j}\}_{j=1}^J$, Target base $\mathbf{h}_{tgt}^{(0)}$, Prior $\mathbf{s}_{comp}$}
\KwOut{Contextual representation $\mathbf{e}_{inter}$}

\For{$j \leftarrow 1$ to $J$}{
    $\mathbf{h}_j^{(0)} \leftarrow \text{TextAnchored\_MHA}(\mathbf{h}_{T,j}^c, [\mathbf{H}_{A,j} \parallel \mathbf{H}_{V,j}])$\;
}
Inject tension prior: $\tilde{\mathbf{h}}_{tgt}^{(0)} \leftarrow \mathbf{h}_{tgt}^{(0)} + \text{LayerNorm}(\mathbf{W}_{pri} \mathbf{s}_{comp})$\;
Initialize conversational topology $\mathbf{H}^{(0)} \leftarrow [\mathbf{h}_1^{(0)}; \dots ; \mathbf{h}_J^{(0)} ; \tilde{\mathbf{h}}_{tgt}^{(0)}]$\; 

\For{layer $k \leftarrow 1$ to $K_{gnn}$}{
    Update node polarities $\mathbf{p}_n \leftarrow \text{L2Norm}(\text{MLP}_{pol}^{ctx}(\mathbf{H}_n^{(k-1)}))$\;
    \For{each relation type $r \in \{seq, ctx, spk\}$}{
        Calculate contradiction penalties $C_{ij} \leftarrow 1 - \mathbf{p}_i \cdot \mathbf{p}_j$\;
        Compute topology-masked attention $\alpha_{ij,r}$ modulated by $C_{ij}$ and $\alpha_{ctx}$\;
        $\mathbf{H}_i^{(k)} \leftarrow \text{ReLU}\left(\sum_{j} \alpha_{ij,r} \mathbf{W}_r \mathbf{H}_j^{(k-1)}\right)$\;
    }
}

\Return $\mathbf{e}_{inter} \leftarrow \mathbf{H}_{J+1}^{(K_{gnn})}$\;
\end{algorithm}

\subsection{Step 5: Adaptive Fusion and Optimization}
\label{sec:fusion_and_loss}
To address heterogeneity in sarcasm, we fuse the atomic-level micro-conflict vector $\mathbf{e}_{atomic}$ with the contextual representation $\mathbf{e}_{inter}$. Strictly excluding branch-specific graph features from fusion prevents semantic looping. Bridging scale discrepancies via a learnable scaling factor, independent GELU and LayerNorm projections form a dual-granularity matrix $\mathbf{M}_{fuse} = [\text{Proj}(\mathbf{e}_{atomic}); \text{Proj}(\mathbf{e}_{inter})]$. A parameterized scorer dynamically routes the critical incongruity signal via attention weights $\mathbf{a}_{fuse} = \text{Softmax}(\mathbf{w}_{fuse}^\top \tanh(\mathbf{M}_{fuse} \mathbf{W}_F))$, yielding the final prediction $\hat{y} = \text{Classifier}(\mathbf{M}_{fuse}^\top \mathbf{a}_{fuse})$.

Standard cross-entropy $\mathcal{L}_{cls}$ drives binary classification, while an inconsistency-aware supervised contrastive loss $\mathcal{L}_{con}$ explicitly structures the representation space, forcing direct comparison of intrinsic structural conflicts. Projecting concatenated branch-specific cross-modal differences via an MLP yields an L2-normalized inconsistency representation $\mathbf{z}_{incon}^{(i)}$. For anchor $i$ in batch $\mathcal{B}$, the contrastive loss clusters positive samples of the same class while pushing away all other valid samples:
\begin{equation}\small \small
\mathcal{L}_{con} = \frac{1}{|\mathcal{B}|}\sum_{i \in \mathcal{B}} \frac{-1}{|\mathcal{P}(i)|} \sum_{p \in \mathcal{P}(i)} \log \frac{\exp(\mathbf{z}_{incon}^{(i)} \cdot \mathbf{z}_{incon}^{(p)} / \tau)}{\sum_{a \in \mathcal{A}(i)} \exp(\mathbf{z}_{incon}^{(i)} \cdot \mathbf{z}_{incon}^{(a)} / \tau)},
\label{eq:con_loss}
\end{equation}
where $\mathcal{P}(i)$ is the positive set and $\mathcal{A}(i)$ encompasses all valid samples excluding the anchor. To prevent mapping noise due to premature subclass separation, this contrastive objective aligns along a generalized incongruity manifold. Abstracting commonalities of sarcasm into a unified positive cluster anchors robust boundaries amid noisy signals, preventing semantic collapse.

The two-stage asymmetric strategy optimizes the network. The initial cold-start phase anchors the polarity space by adding an MSE loss $\mathcal{L}_{valence}$ computing the divergence between native continuous valence scores and a Tanh-bounded polarity probe output, preventing gradient explosion:
\begin{equation}\small 
    \mathcal{L}_{total} = \lambda_{cls} \mathcal{L}_{cls} + \lambda_{con} \mathcal{L}_{con} + \lambda_{val} \mathcal{L}_{valence}.
\label{eq:total_loss_warmup}
\end{equation}
Subsequently, detaching the auxiliary valence supervision transitions the network into topological refinement, relying solely on cross-entropy and contrastive losses:
\begin{equation}\small 
    \mathcal{L}_{total} = \lambda_{cls} \mathcal{L}_{cls} + \lambda_{con} \mathcal{L}_{con}.
\label{eq:total_loss_final}
\end{equation}
Weighting the contrastive loss heavily asserts that segregating conflict topologies remains fundamentally more critical than naive label-mapping gradients.


\section{Experimental Settings}
\label{sec:experiments}

This section details the datasets, baseline models, and implementation details.

\subsection{Datasets}
\label{sec:datasets_detail}

We use \texttt{MUStARD} \cite{ref3} as our foundational benchmark, a dataset specifically compiled for multimodal sarcasm detection, drawn from clips of popular television shows. To assess the model's structural robustness and pragmatic reasoning capabilities, we extend the evaluation to two progressively challenging datasets. The first, \texttt{MUStARD++} \cite{ref_mustard_plus}, introduces fine-grained annotations and corrects surface labels. The second is \texttt{\texttt{MUStARD++} Balanced} \cite{zhang2025sarcasmbench}. All datasets contain text, audio, and video modalities. Following standard affective computing protocols to prevent the memorization of specific facial and vocal features, we evaluate all models using a 5-fold cross-validation split.
\begin{table*}[!b]
\centering
\footnotesize
\caption{Comprehensive performance comparison on the \texttt{MUStARD} benchmark. Bold indicates best performance; \underline{underline} indicates second-best.}
\label{tab:main_results}
\setlength{\tabcolsep}{7.3mm}
\begin{tabular}{l l cccc}
\toprule
\textbf{Category} & \textbf{Model} & \textbf{Prec (\%)} & \textbf{Rec (\%)} & \textbf{F1 (\%)} & \textbf{Acc (\%)} \\
\midrule
\multirow{2}{*}{\textit{Text-Only}}
 & SVM-TK \cite{ref3}  & 69.40 & 71.20 & 70.30 & 72.60 \\
 & BERT \cite{devlin2019bert}  & 65.60 & 64.30 & 64.70 & 64.30 \\
\midrule
\multirow{3}{*}{\textit{Tensor Fusion}}
 & TFN \cite{ref7}  & 66.10 & 64.20 & 65.30 & 66.50 \\
 & LMF \cite{eke2021multi}  & 68.30 & 67.10 & 67.70 & 68.20 \\
 & IKM \cite{yue2023knowlenet}  & 70.10 & 69.30 & 69.70 & 70.50 \\
\midrule
\multirow{3}{*}{\textit{Cross-Modal Attention}}
 & MAG-BERT \cite{ref38}  & 69.40 & 69.50 & 69.40 & 69.40 \\
 & FiLM \cite{ref_film}  & 67.30 & 66.20 & 66.70 & 67.00 \\
 & COSMOS \cite{ouyang2022cosmo}  & 70.80 & 69.10 & 69.90 & 70.20 \\
\midrule
\multirow{5}{*}{\textit{Graph-Based}}
 & ADG \cite{lou2021affective}  & 72.10 & 71.50 & 71.80 & 72.00 \\
 & GCN-DN \cite{liang2022multimodal}  & 73.80 & 72.20 & 73.00 & 73.50 \\
 & G$^2$SAM \cite{ref_aaai24_graph}  & 75.30 & 73.60 & 73.50 & 73.60 \\
 & GNN-CSO  \cite{singh2024enhancing}  & 71.20 & 70.80 & 71.00 & 71.50 \\
 & CMGCN \cite{liang2021cmgcn}  & 74.92 & 72.25 & 71.58 & 72.37 \\
\midrule
\multirow{6}{*}{\textit{Advanced Multitask}}
 & ICON \cite{hazarika2018icon}  & 71.50 & 70.20 & 70.80 & 71.20 \\
 & QPM  \cite{liu2021does}  & 77.50 & 77.60 & 77.50 & 77.50 \\
 & MO-Sarcation \cite{tomar2023your}  & 77.90 & 77.90 & 77.90 & 77.90 \\
 & CESDN \cite{li2024attention}  & 76.20 & 74.20 & 75.20 & 75.50 \\
 & VyAnG-Net  \cite{pandey2025vyang}  & \underline{78.80} & \underline{78.20} & \underline{78.50} & \underline{79.90} \\
 & ESAM  \cite{ESAM}  & 71.95 & 68.97 & 69.05 & 71.05 \\
\midrule
\multirow{3}{*}{\textit{Causal \& Debiasing}}
 & T-FCD \cite{zhu2024tfcd}  & 73.50 & 72.80 & 73.10 & 73.40 \\
 & CL-Debias \cite{qiao2024debiasing}  & 74.10 & 73.50 & 73.80 & 74.00 \\
 & MVIL  \cite{guo2025multi}  & 75.80 & 74.90 & 75.30 & 75.60 \\
\midrule
\multirow{3}{*}{\textit{Large Multimodal Models}}
 & GPT-4o \cite{yao2025sarcasm}  & 71.82 & 68.10 & 69.91 & 70.69 \\
 & Llama 3-8B \cite{yao2025sarcasm} & 67.29 & 51.05 & 61.26 & 68.90 \\
 & Qwen 2-7B \cite{yao2025sarcasm} & 70.98 & 54.35 & 61.46 & 65.94 \\
\midrule
\multirow{4}{*}{\textbf{Ours}} 
 & PC-MNet (V-Only) & 63.40 & 62.10 & 62.74 & 63.50 \\
 & PC-MNet (A-Only) & 68.50 & 67.30 & 67.89 & 68.20 \\
 & PC-MNet (T-Only) & 75.40 & 74.80 & 75.09 & 75.80 \\
 \cmidrule{2-6}
 & \textbf{PC-MNet (Full)} & \textbf{83.61} & \textbf{81.09} & \textbf{81.64} & \textbf{82.46} \\
\bottomrule
\end{tabular}
\end{table*}
\subsection{Baseline Models}
To rigorously evaluate the proposed PC-MNet, we comprehensively compare our approach against seven distinct categories of baselines, ranging from unimodal architectures to foundation models. The first category encompasses text-only models, specifically SVM-TK \cite{ref3} and BERT \cite{devlin2019bert}, which serve as foundational baselines that rely entirely on linguistic cues to detect sarcasm. Recognizing the necessity of non-verbal information, the second category includes tensor fusion approaches such as TFN \cite{ref7}, LMF \cite{eke2021multi}, and IKM \cite{yue2023knowlenet}. These models explicitly calculate outer products or integrate external knowledge to capture joint multimodal distributions. Subsequently, we evaluate cross-modal attention mechanisms, including MAG-BERT \cite{ref38}, FiLM \cite{ref_film}, and COSMOS \cite{ouyang2022cosmo}, which leverage sophisticated alignment strategies and feature modulation to capture inter-modal dependencies and contextual grounding.

Furthermore, given the topological nature of conversational dynamics, we benchmark against graph-based models, including ADG \cite{lou2021affective}, GCN-DN \cite{liang2022multimodal}, G$^2$SAM \cite{ref_aaai24_graph}, CMGCN \cite{liang2021cmgcn}, and GNN-CSO \cite{singh2024enhancing}. These architectures construct heterogeneous topological structures to explicitly model long-range contextual dependencies and cross-modal relational features. Additionally, advanced multitask frameworks including ICON \cite{hazarika2018icon}, QPM \cite{liu2021does}, MO-Sarcation \cite{tomar2023your}, CESDN \cite{li2024attention}, VyAnG-Net \cite{pandey2025vyang}, and ESAM \cite{ESAM} are included. These frameworks introduce auxiliary objectives, such as sentiment classification or sentiment constraints, to provide complementary supervisory signals for sarcasm detection. To address the critical issue of dataset artifacts, we also compare with causal debiasing methods, namely T-FCD \cite{zhu2024tfcd}, CL-Debias \cite{qiao2024debiasing}, and MVIL \cite{guo2025multi}, which mitigate spurious correlations through counterfactual reasoning or debiased contrastive learning. Notably, we explicitly reproduced the recent ESAM \cite{ESAM} model, yielding an overall accuracy of 71.05\% and a Macro-F1 score of 69.05\% under our experimental setting.

\subsection{Implementation Details}
In our experiments, we use BERT-large and Wav2Vec 2.0-base to extract textual and acoustic features, respectively. For the visual modality, we combine YOLOv8 with CLIP ViT-B/32. We set the shared encoding dimension to $d_{enc} = 512$ and the polarity space dimension to $d_p = 16$. We configure the heterogeneous graph with $L_{mac} = 2$ layers and set the historical context window size to $J = 3$. The contextual relational penalty is initialized at $\alpha_{ctx} = 0.1$. For the two-stage asymmetric optimization, we set the valence supervision weight $\lambda_{val} = 1.0$ during the first $E_{warm} = 5$ warm-up epochs. Subsequently, we remove the auxiliary supervision and transition the loss weights to $\lambda_{cls} = 0.2$ and $\lambda_{con} = 0.8$ for topological refinement. We train the network on an NVIDIA RTX 4090 GPU for 15 epochs with early stopping. During training, we use the AdamW optimizer with a learning rate of $5 \times 10^{-5}$ and a batch size of 16. Following previous work \cite{guo2025multi}, we adopt Accuracy, Precision, Recall, and F1-score as our evaluation metrics.

\section{Results and Analysis}
\label{sec:main_results}
This section presents the main results, ablation studies, qualitative visualizations, and a case study.

\subsection{Main Results}

Table \ref{tab:main_results} evaluates PC-MNet against 25 baselines across seven paradigms on the \texttt{MUStARD} benchmark. PC-MNet achieves a Macro-F1 score of 81.64\% and an accuracy of 82.46\%, outperforming all baselines. A paired t-test confirms that the 3.14\% absolute F1-score improvement over the strongest multimodal baseline, VyAnG-Net \cite{pandey2025vyang}, is statistically significant ($p < 0.01$). Single-modality variants highlight the necessity of multimodal integration: the text-only baseline achieves a 75.09\% F1-score, while isolated video and audio models struggle with pragmatic ambiguity, yielding 62.74\% and 67.89\%, respectively. Integrating these modalities into PC-MNet yields a 6.55\% absolute F1-score increase over the text-only model. Consequently, this substantial performance gain demonstrates that explicitly modeling cross-modal incongruity, rather than merely pursuing modality similarity, is essential for robust sarcasm detection. Furthermore, an analysis of the advanced baselines reveals distinct methodological limitations. The recent graph-based model CMGCN \cite{liang2021cmgcn} achieves an F1-score of 71.58\%. This result indicates that while constructing cross-modal graphs is beneficial, monolithically entangling modalities without explicit polarity-driven routing inevitably leads to semantic over-smoothing. In contrast, the performance of the ESAM \cite{ESAM} architecture plummets to an F1-score of 69.05\%. This sharp performance decline indicates that the associated multi-task framework, which integrates sentiment constraints, remains effective only in scenarios involving short-form multimodal data devoid of temporal context. However, the substantial performance gap between ESAM and PC-MNet demonstrates that soft sentimental regularization is insufficient without structurally decoupling atomic and contextual incongruities. Additionally, large multimodal models such as Llama 3-8B \cite{yao2025sarcasm} exhibit a severe precision-recall imbalance on this task, achieving a precision of 67.29\% but suffering a significant drop in recall to 51.05\%. This significant disparity indicates a reliance on global textual surface cues rather than fine-grained multimodal reasoning. Conversely, PC-MNet maintains a stable balance, achieving 83.61\% Macro-Precision and 81.09\% Macro-Recall. Consequently, these comparative results confirm that for fine-grained incongruity verification, explicit structural constraints are fundamentally more effective than unconstrained parameter scaling. To further ensure the statistical reliability of these findings, all reported metrics for PC-MNet are averaged over five independent runs with different random seeds. The minimal variance observed rigorously validates that the performance gains stem from the proposed architectural design rather than favorable random initializations, thereby corroborating the aforementioned $p < 0.01$ significance level.

\begin{table}[!t]
\footnotesize
\centering
\caption{Ablation study on the \texttt{MUStARD} benchmark.}
\label{tab:ablation}
\setlength{\tabcolsep}{1.35mm}
\begin{tabular}{lccc} 
\toprule
\textbf{Model}  & \textbf{Prec (\%)} & \textbf{Rec (\%)} & \textbf{F1 (\%)} \\
\midrule
\textbf{PC-MNet (Full)}  & \textbf{83.61} & \textbf{81.09} & \textbf{81.64} \\
\midrule
w/o Polarity Modulation  & 77.80 & 75.30 & 76.32 \\
w/o $\mathbf{e}_{atomic}$  & 78.50 & 76.10 & 77.15 \\
w/o $\mathbf{e}_{inter}$  & 79.10 & 76.80 & 77.80 \\
\midrule
\textbf{w/ Tripartite Graph (T-A-V)} & 78.90 & 76.40 & 77.45 \\
\textbf{w/ Direct Fusion of Branch Topologies}  & \textbf{78.10} & \textbf{75.80} & \textbf{76.85} \\
\midrule
w/o $\mathcal{L}_{valence}$  & 79.50 & 77.10 & 78.12 \\
w/o $\mathcal{L}_{con}$ & 76.80 & 74.50 & 75.40 \\
\bottomrule
\end{tabular}
\end{table}

\subsection{Ablation Study}
\label{sec:ablation}
Table \ref{tab:ablation} demonstrates PC-MNet performance stems from a decoupled multi-granularity design. All ablated models underperform the best baseline, VyAnG-Net, validating the fairness of the comparative setup. PC-MNet prevents high-dimensional branch-specific composition features ($\mathbf{h}_{T(V)}^{comp}$ and $\mathbf{h}_{T(A)}^{comp}$) from directly entering the final classifier. Bypassing the scalar congruity routing mechanism via direct fusion results in a 4.79\% drop in F1-score (to 76.85\%). Substituting parallel bipartite-dominant graphs with a single fully connected tripartite graph reduces the F1-score to 77.45\%. The degradation demonstrates that projecting unaligned multimodal signals onto a monolithic graph inevitably causes semantic over-smoothing. Replacing polarity-modulated attention with standard dot-product attention causes a 5.32\% F1-score drop (to 76.32\%). The decrease highlights that explicitly modeling cross-modal contradictions over similarities remains critical for effective sarcasm detection. Removing the micro-level atomic module or inter-sentence contextual GNN drops the F1-score by 4.49\% and 3.84\%, respectively, validating the necessity of dual-granularity integration. Relying solely on cross-entropy loss causes a 6.24\% F1-score degradation. The decline indicates the necessity of a contrastive loss ($\mathcal{L}_{con}$) to explicitly cluster inconsistent topologies. Discarding the initial 5-epoch valence warm-up phase drops F1-score by 3.52\%. Lacking continuous valence loss to anchor initial polarity spaces destabilizes the network, hindering subsequent contrastive learning.

\begin{figure}[!t]
\centering
    \subfigure[Without $\mathcal{L}_{con}$]{
      \includegraphics[width=0.48\columnwidth]{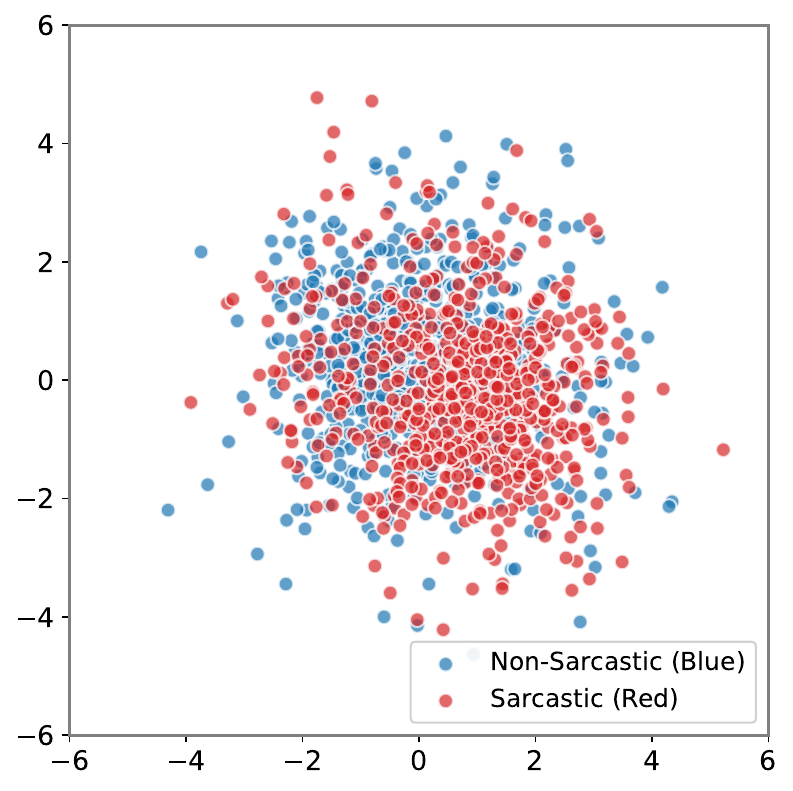}\label{fig:tsne_wo}}
    \subfigure[Full PC-MNet]{\includegraphics[width=0.48\columnwidth]{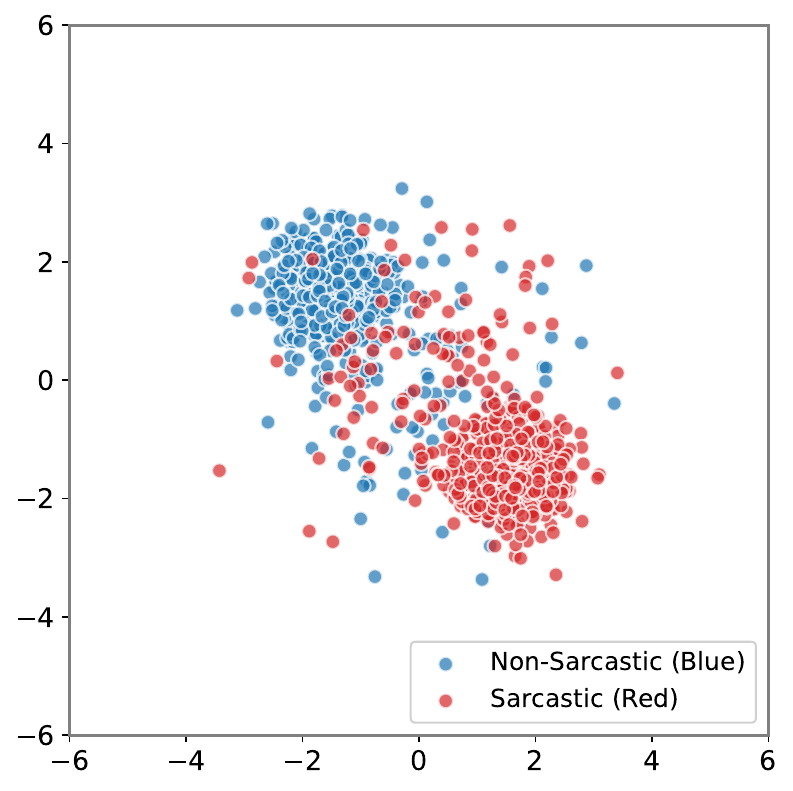}\label{fig:tsne_full}}
\caption{Visualization of Feature Distributions for Sincere and Sarcastic Samples Without $\mathcal{L}_{con}$ and With Full PC-MNet.}
\label{fig:tsne}
\end{figure}

\begin{figure}[!b]
\centering
\subfigure[standard attention heatmap.]
    {
        \includegraphics[width=0.325\textwidth]{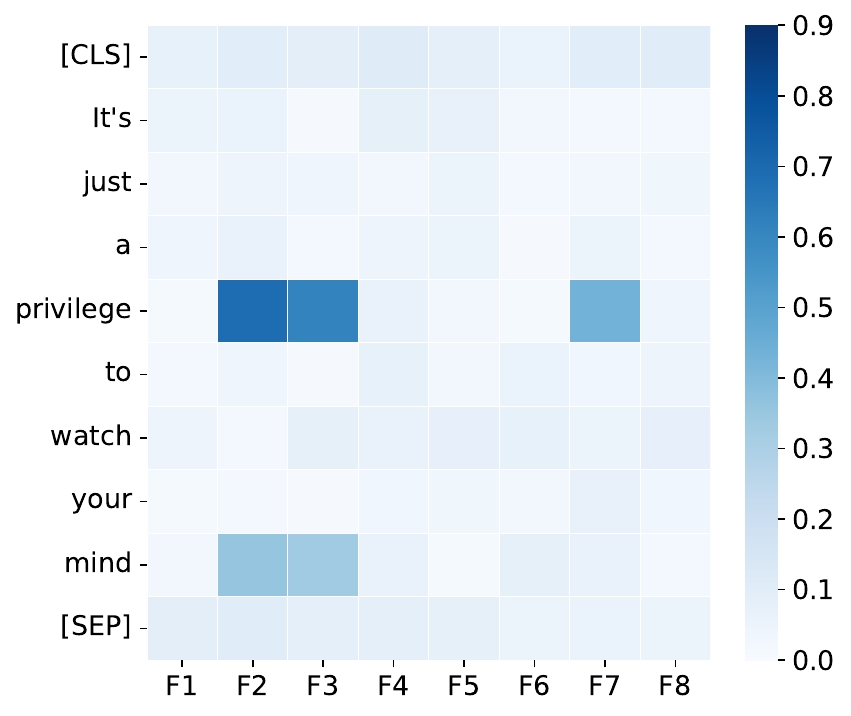}
    }
\subfigure[Polarity Modulated Attention Heatmap.]
    {
        \includegraphics[width=0.325\textwidth]{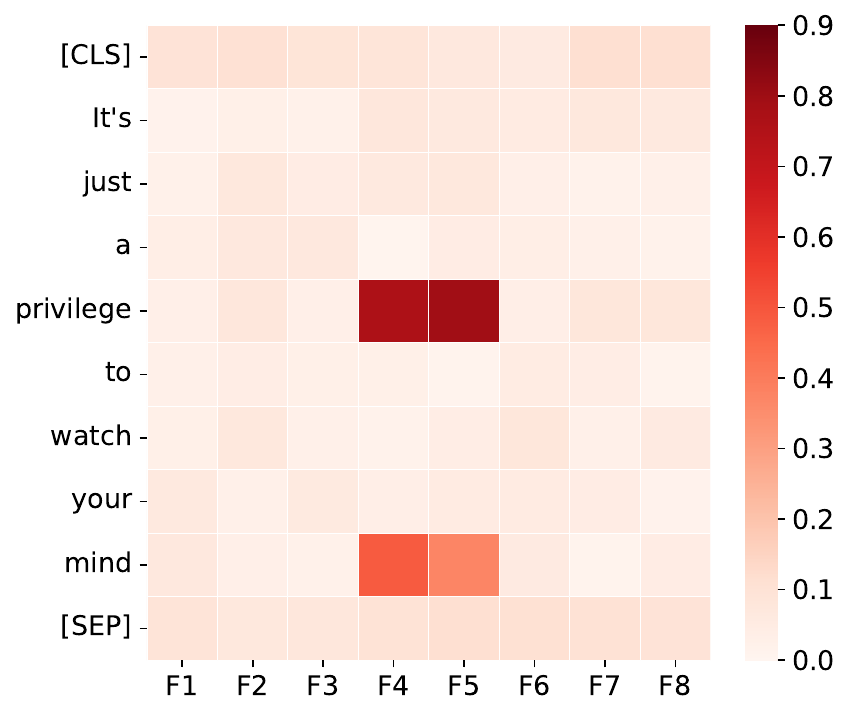}
    }
\caption{Visualization of Attention Heatmaps for Standard Similarity-Seeking and Polarity-Modulated Contradiction-Seeking Mechanisms.}
\label{fig:attention_heat}
\end{figure}

\subsection{Visualization Discussion}
\label{sec:qualitative_vis}
\begin{figure}[!t]
\centering
\includegraphics[width=0.48\textwidth]{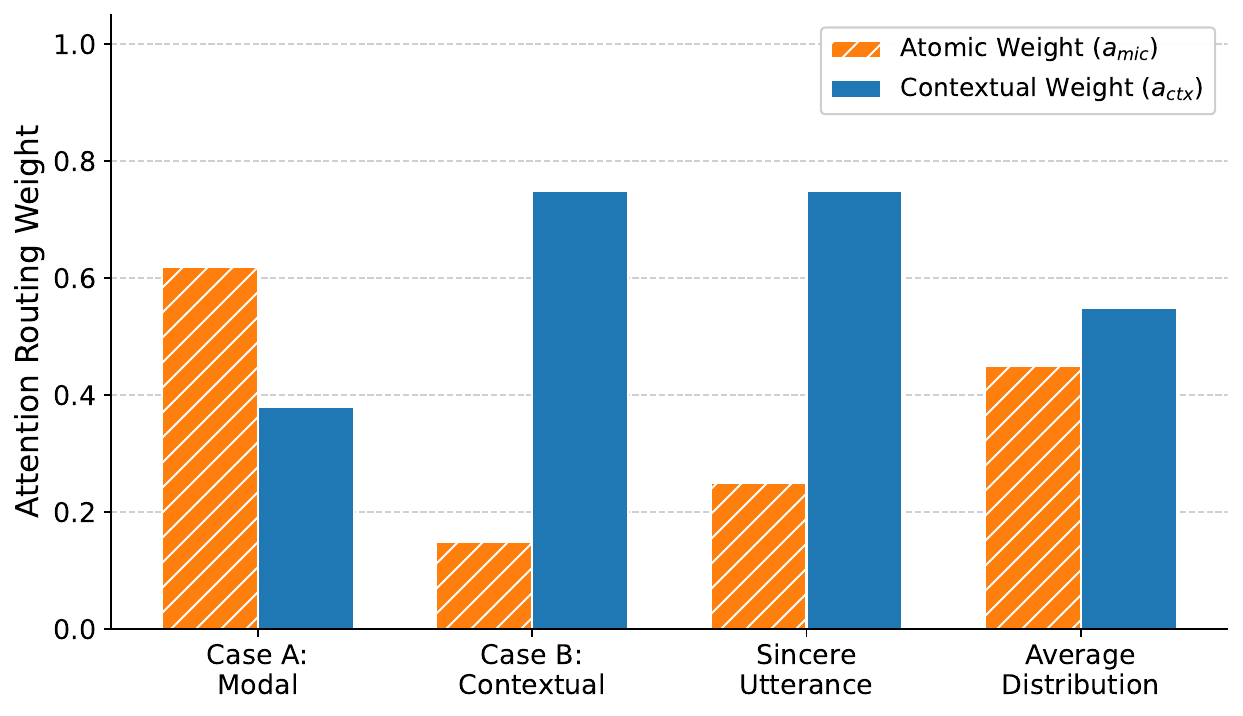}
\caption{Dynamic routing weight distribution ($\mathbf{a}_{fuse}$) across different pragmatic scenarios.}
\label{fig:routing_weights}
\end{figure}

Visualizations can validate our architectural design. Figure \ref{fig:tsne} plots t-SNE projected test representations ($\mathbf{e}_{final}$). Without contrastive loss $\mathcal{L}_{con}$ (Figure \ref{fig:tsne_wo}), standard optimization causes substantial sarcastic-sincere overlap (0.68 purity). Conversely, the complete PC-MNet (Figure \ref{fig:tsne_full}) forms highly separable clusters (0.81 purity). This distinct separation demonstrates contrastive optimization effectively isolates latent sarcastic representations. Figure \ref{fig:attention_heat} visualizes cross-attention weights for an utterance pairing a mid-sentence eye-roll (frames F4 and F5) with the positive verbal cue privilege. Standard attention incorrectly highlights smiling frames (F2 and F3). In contrast, polarity-modulated attention accurately localizes contradictory frames. This precise localization confirms that explicitly modeling contradictions captures transient micro-level incongruities essential for robust sarcasm detection. Finally, Figure \ref{fig:routing_weights} tracks dynamic routing weights $\mathbf{a}_{fuse} = [a_{mic}, a_{ctx}]^\top$. The tracking verifies that the fusion layer adaptively shifts computational focus based on underlying sarcasm mechanisms, avoiding feature redundancy and ensuring optimal dual-level integration.

\begin{figure*}[!b]
    \centering
    \subfigure[Atomic Modulation Weight ($\alpha$).]
    {
        \includegraphics[width=0.315\textwidth]{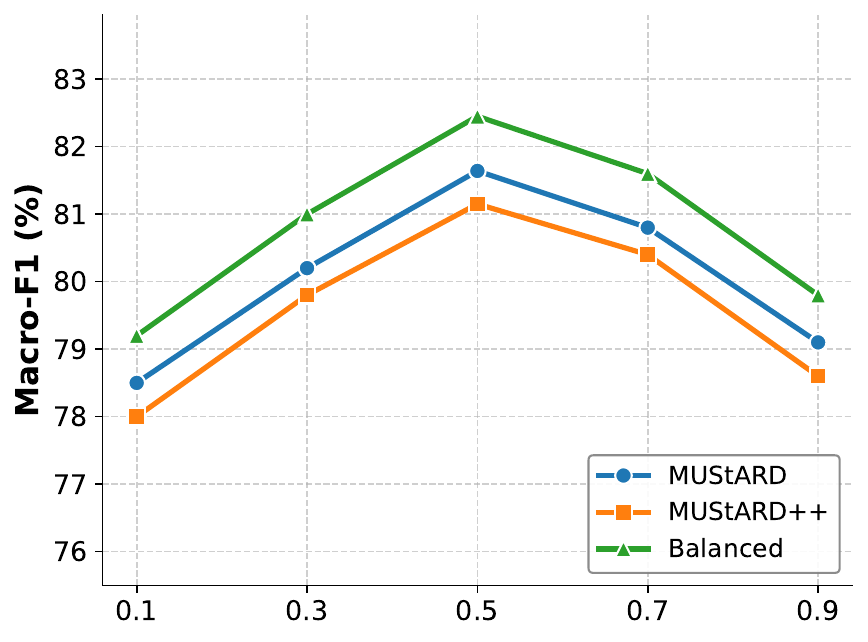}
        \label{Fig:hyper_alpha}
    }
    \subfigure[Graph Depth ($L_{mac}$).]
    {
        \includegraphics[width=0.315\textwidth]{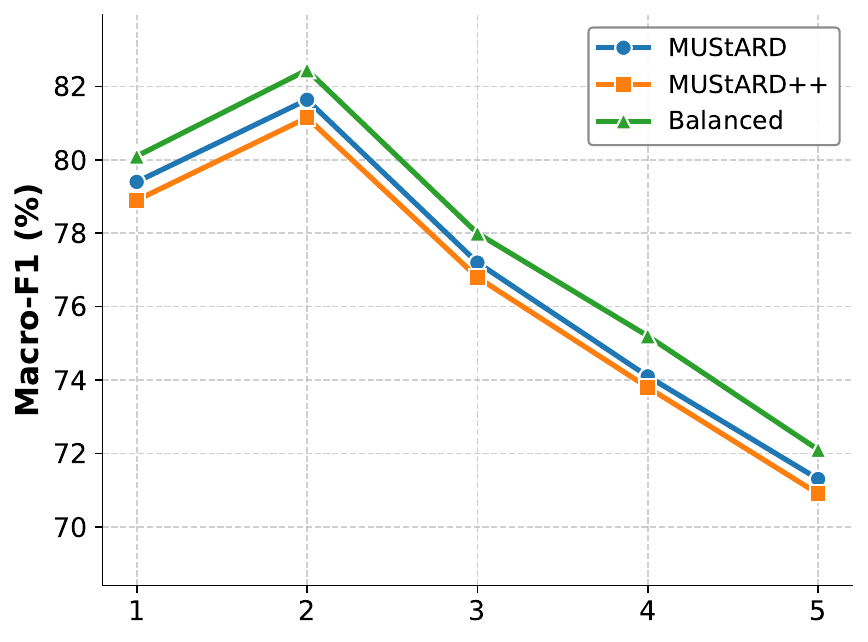}
        \label{Fig:hyper_gnn}
    }
    \subfigure[Contrastive Weight ($\lambda_{con}$).]
    {
        \includegraphics[width=0.315\textwidth]{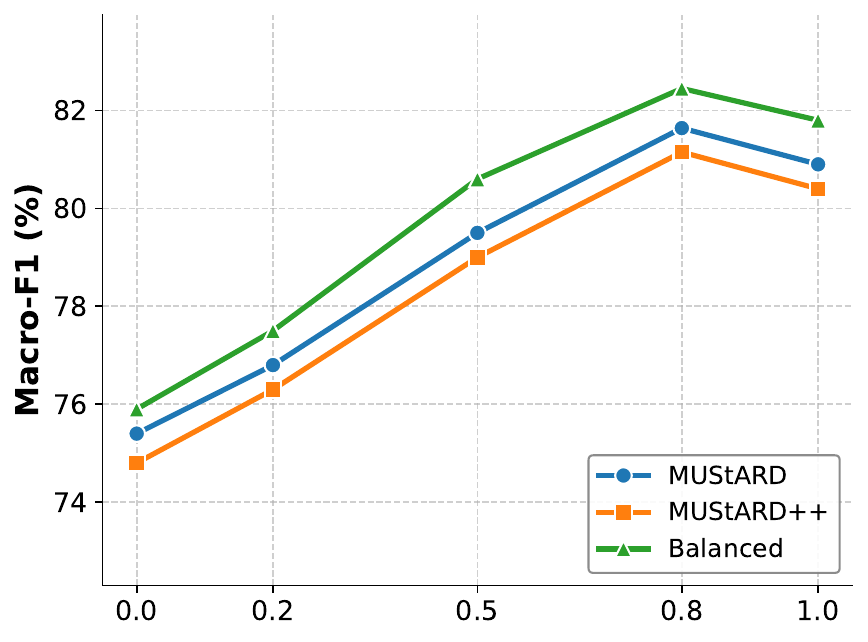}
        \label{Fig:hyper_loss}
    }
    \subfigure[Contrastive Temperature ($\tau$).]
    {
        \includegraphics[width=0.315\textwidth]{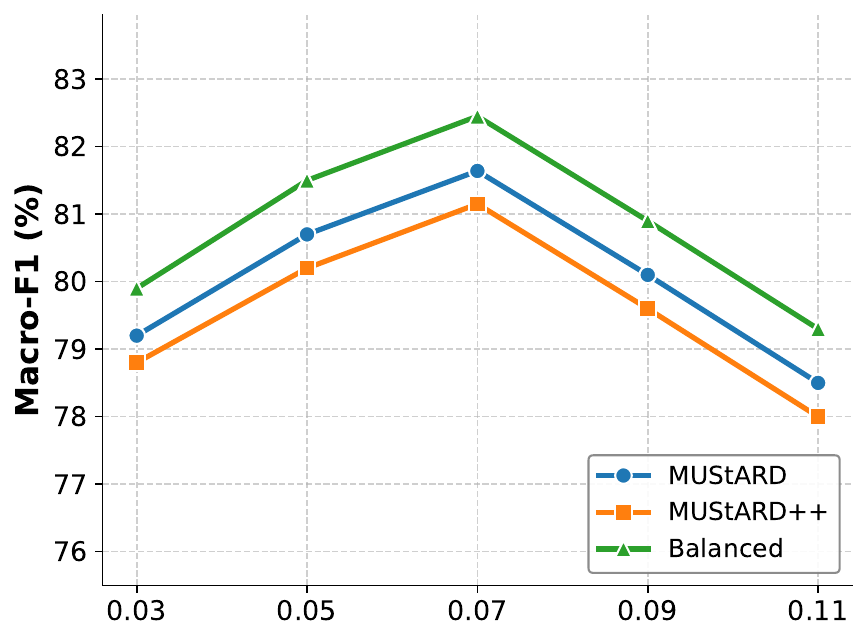}
        \label{Fig:hyper_tau}
    }
    \subfigure[Contextual Penalty ($\alpha_{ctx}$).]
    {
        \includegraphics[width=0.315\textwidth]{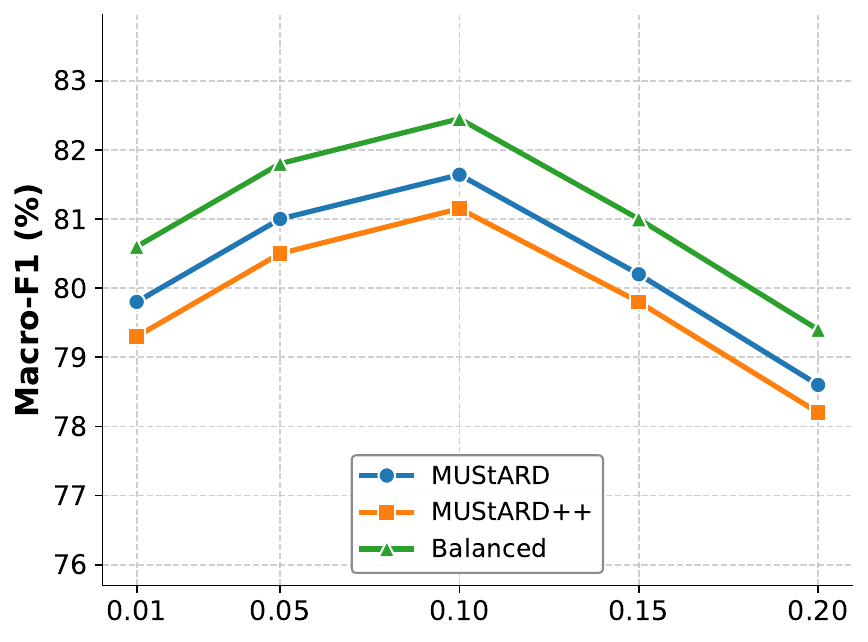}
        \label{Fig:hyper_actx}
    }
    \subfigure[Warm-up Epochs ($E_{warm}$).]
    {
        \includegraphics[width=0.315\textwidth]{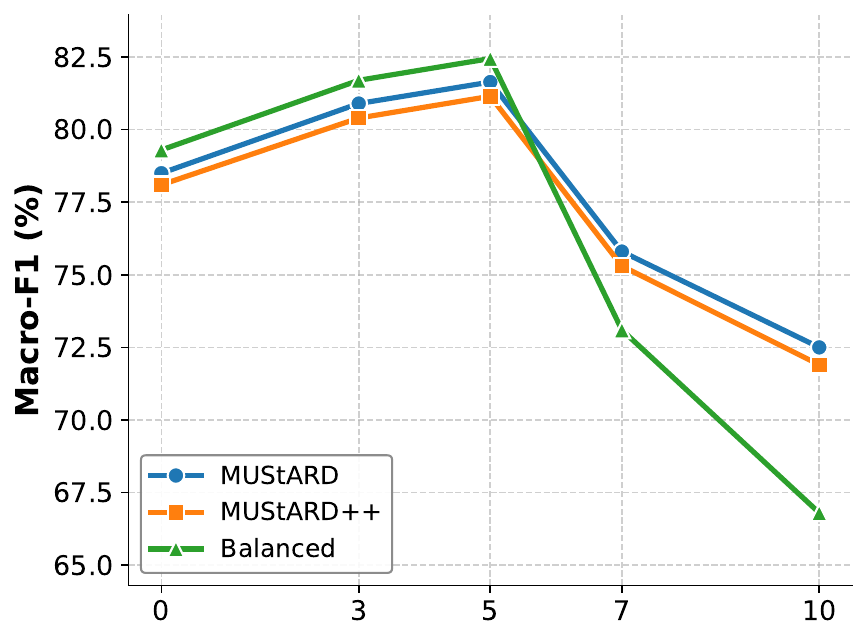}
        \label{Fig:hyper_ewarm}
    }
    \caption{Comprehensive hyperparameter sensitivity analysis tracking F1 Score trajectories across three dataset variants.}
    \label{Fig:hyperparams}
\end{figure*}

\subsection{Case Study}

\textbf{Success Case A -- Modal Incongruity:} For an utterance delivered with an eye-roll, the textual sentiment is positive while the visual polarity is negative. Our detection pipeline isolates the facial expression. Because micro-level conflicts dominate, the network appropriately assigns the highest routing weight to the atomic-level representation.

\textbf{Success Case B -- Contextual Incongruity:} A speaker says, ``\textit{I don't think I'll be able to stop thinking about it}'' in a deadpan tone. Identifying no micro-level conflicts, PC-MNet shifts focus to the inter-sentence GNN, establishing a contradiction edge with the speaker's historical disinterest. Assigning a contextual weight of 0.85 and a micro weight of 0.15, this dynamic delegation validates the dual-granularity strategy.

\textbf{Error Analysis -- Case C:} A character mocks a messy room using pseudo-scientific vocabulary in a sincere tone. PC-MNet incorrectly predicts this as non-sarcastic. When sarcasm relies entirely on semantic irony without observable multimodal affective shifts, the model defaults to a literal interpretation.

\subsection{Cross-Dataset Generalization}

Models exploiting dataset imbalances often suffer severe performance degradation when evaluated on uniform distributions. 
To formalize the evaluation protocol, the proposed framework and baselines are assessed across the original \texttt{MUStARD} dataset and its two extended datasets (i.e., \texttt{MUStARD++} and \texttt{\texttt{MUStARD++} Balanced}). 
The purpose is to systematically test architectural robustness against increasing data scale and forced label parity. Table \ref{tab:distribution_robustness} summarizes the overall generalization trends across these datasets. As demonstrated in Table \ref{tab:distribution_robustness}, a distinct performance divergence emerges under the rigorous constraints of \texttt{\texttt{MUStARD++} Balanced} dataset. Specifically, baseline performances decay significantly. For instance, massive unconstrained models such as Llama 3-8B suffer severe degradation, plummeting to F1-scores of 63.80\% and 61.10\%, respectively. This specific degradation occurs because these baselines rely heavily on spurious correlations and class imbalances within the training data, rather than on fine-grained pragmatic reasoning. Conversely, the performance of PC-MNet remains remarkably stable, even increasing to an 82.45\% F1-score on the balanced set. This robust improvement validates a core architectural advantage: the proposed framework relies entirely on a universal, generalized incongruity manifold. Consequently, the PC-MNet architecture effectively resists dataset biases and maintains high detection efficacy across varying distributions.

\begin{table}[!t]
\centering
\caption{Cross-Dataset Generalization (Macro-F1 \%). Bold indicates best; \underline{underline} indicates second-best.}
\label{tab:distribution_robustness}
\setlength{\tabcolsep}{2.5mm}
\begin{tabular}{l c c c }
\toprule
\textbf{Model} & \textbf{MUStARD} & \textbf{\texttt{MUStARD++}} & \begin{tabular}{@{}c@{}}\textbf{\texttt{MUStARD++}} \\ \textbf{Balanced}\end{tabular} \\
\midrule
G$^2$SAM         & 73.50 & 71.24 & 70.85 \\
MVIL             & 75.30 & 74.15 & 73.90 \\
ESAM             & 69.05 & 65.22 & 69.05 \\
VyAnG-Net        & \underline{78.50} & 76.92 & 76.10 \\
MO-Sarcation     & 77.90 & \underline{77.80} & \underline{76.55} \\
Llama 3-8B        & 61.26 & 63.80 & 61.10 \\
\midrule
\textbf{PC-MNet (Ours)}           & \textbf{81.64} & \textbf{81.15} & \textbf{82.45} \\
\bottomrule
\end{tabular}
\end{table}

\subsection{Hyperparameter Analysis}\label{sec:hyper_analysis}Figure \ref{Fig:hyperparams} analyzes six critical hyperparameters, confirming the selected configurations avoid overfitting specific data distributions. Detection performance peaks at a micro-atomic modulation weight of $\alpha_{mic}=0.5$, proving polarity modulation should enhance, not override, base semantic features. Setting composition graph depth to $L_{mac}=2$ achieves optimal performance, whereas deeper networks ($L_{mac} \ge 3$) suffer severe semantic over-smoothing, blurring localized incongruities. Additionally, a contextual penalty of $\alpha_{ctx}=0.1$ yields peak performance. The specific threshold maintains foundational dialogue context while amplifying necessary pragmatic shifts. Regarding optimization dynamics, Figure \ref{Fig:hyper_loss} shows that a contrastive weight of $\lambda_{con}=0.8$ yields the highest F1-score. The performance peak rigorously validates the asymmetrical loss strategy, prioritizing embedding-space topology structuring. Moreover, a contrastive temperature of $\tau=0.07$ effectively calibrates the negative penalty during contrastive learning. Finally, Figure \ref{Fig:hyper_ewarm} demonstrates that extending auxiliary valence supervision beyond 5 epochs significantly degrades model performance, dropping the \texttt{\texttt{MUStARD++} Balanced} dataset F1-score to 66.80\%. The substantial degradation suggests prolonged exposure to explicit scalar labels overly restricts the representation space, underscoring the necessity of the proposed two-stage optimization strategy.

\section{Conclusion}
\label{sec:conclusion}

We propose PC-MNet, a hierarchical multi-granularity framework for multimodal sarcasm detection explicitly modeling pragmatic incongruities over simple feature fusion. Polarity-modulated attention captures fine-grained cross-modal conflicts, while parallel bipartite-dominant heterogeneous graphs extract intra-utterance structural sarcasm. A strict scalar congruity routing mechanism blocks high-dimensional features from the classifier, routing them as scalar-congruity priors for contextual GNNs and anchors for inconsistency-aware contrastive loss, averting late-fusion gradient interference. Experiments on \texttt{MUStARD} and its balanced datasets demonstrate consistent state-of-the-art performance. Ablations confirm dual-granularity routing and two-stage optimization, utilizing valence warm-up to stabilize representations before contrastive refinement. Relying on observable cues, the model occasionally struggles with dry irony, requiring extensive world knowledge. Future work integrates Multimodal Large Language Models (MLLMs), injecting commonsense priors into nodes, extending to streaming multi-party dialogues. Ultimately, PC-MNet provides a robust paradigm for modeling cross-modal contradiction over mere similarity, advancing broader affective computing tasks.
\balance
\bibliographystyle{IEEEtran}
\bibliography{references}

@inproceedings{ref3,
  title={Towards multimodal sarcasm detection},
  author={Castro, Santiago and Hazarika, Devamanyu and P{\'e}rez-Rosas, Ver{\'o}nica and Zimmermann, Roger and Mihalcea, Rada and Poria, Soujanya},
  booktitle={Proc. 57th Annu. Meeting Assoc. Comput. Linguist.},
  pages={4619--4629},
  year={2019}
}

@inproceedings{ghosh2017contextual,
  title={The role of conversation context for sarcasm detection in online interactions},
  author={Ghosh, Debanjan and Fabbri, Alexander Richard and Muresan, Smaranda},
  booktitle={Proc. 18th Annu. SIGdial Meeting Discourse Dialogue},
  pages={186--196},
  year={2017}
}

@inproceedings{hasnat2022understanding,
  title={Understanding sarcasm from {Reddit} texts using supervised algorithms},
  author={Hasnat, Fahim and Hasan, Md Mazidul and Nasib, Abdullah Umar and Adnan, Ashik and Khanom, Nazifa and Islam, SM Mahsanul and Mehedi, Md Humaion Kabir and Iqbal, Shadab and Rasel, Annajiat Alim},
  booktitle={Proc. IEEE 10th Region 10 Humanitarian Technol. Conf.},
  pages={1--6},
  year={2022}
}

@inproceedings{guo2025multi,
  title={Multi-view incongruity learning for multimodal sarcasm detection},
  author={Guo, Diandian and Cao, Cong and Yuan, Fangfang and Liu, Yanbing and Zeng, Guangjie and Yu, Xiaoyan and Peng, Hao and Yu, Philip S},
  booktitle={Proc. 31st Int. Conf. on Comput. Linguist.},
  pages={1754--1766},
  year={2025}
}

@inproceedings{qiao2024debiasing,
  title={Debiasing Multimodal Sarcasm Detection with Contrastive Learning},
  author={Jia, Mengzhao and Xie, Can and Jing, Liqiang},
  booktitle={Proc. 38th AAAI Conf. Artif. Intell.},
  volume={38},
  number={16},
  pages={18354--18362},
  year={2024}
}

@inproceedings{pan2024survey,
  title={A Survey of Multimodal Sarcasm Detection},
  author={Farabi, Shafkat  and  Ranasinghe, Tharindu  and  Kanojia, Diptesh  and  Kong, Yu  and  Zampieri, Marcos},
  booktitle={Proc. 33rd Int. Joint Conf. Artif. Intell.},
  pages = {8020--8028},
  year={2024}
}

@article{zhu2023prompt,
  title={Prompt-based learning for unpaired image captioning},
  author={Zhu, Peipei and Wang, Xiao and Zhu, Lin and Sun, Zhenglong and Zheng, Wei-Shi and Wang, Yaowei and Chen, Changwen},
  journal={IEEE Trans. Multimedia},
  volume={26},
  pages={379--393},
  year={2023}
}

@article{wang2022cross,
  title={Cross-modal enhancement network for multimodal sentiment analysis},
  author={Wang, Di and Liu, Shuai and Wang, Quan and Tian, Yumin and He, Lihuo and Gao, Xinbo},
  journal={IEEE Trans. Multimedia},
  volume={25},
  pages={4909--4921},
  year={2022}
}

@article{yang2020image,
  title={Image-text multimodal emotion classification via multi-view attentional network},
  author={Yang, Xiaocui and Feng, Shi and Wang, Daling and Zhang, Yifei},
  journal={IEEE Trans. Multimedia},
  volume={23},
  pages={4014--4026},
  year={2020}
}

@inproceedings{ref_aaai24_graph,
  title={{G\^{} 2SAM}: Graph-Based Global Semantic Awareness Method for Multimodal Sarcasm Detection},
  author={Wei, Yiwei and Yuan, Shaozu and Zhou, Hengyang and Wang, Longbiao and Yan, Zhiling and Yang, Ruosong and Chen, Meng},
  booktitle={Proc. 38th AAAI Conf. Artif. Intell.},
  volume={38},
  pages={9151--9159},
  year={2024}
}

@article{campbell2012necessary,
  title={Are there necessary conditions for inducing a sense of sarcastic irony?},
  author={Campbell, John D and Katz, Albert N},
  journal={Discourse Processes},
  volume={49},
  number={6},
  pages={459--480},
  year={2012}
}

@inproceedings{riloff2013sarcasm,
  title={Sarcasm as contrast between a positive sentiment and negative situation},
  author={Riloff, Ellen and Qadir, Ashequl and Surve, Prafulla and De Silva, Lalindra and Zingano, Nathan and Xia, Yi},
  booktitle={Proc. 2013 Conf. Emp. Methods Natural Lang. Process.},
  pages={704--714},
  year={2013}
}

@inproceedings{tay2018reasoning,
  title={Reasoning with Sarcasm by Reading In-Between},
  author={Tay, Yi and Luu, Anh Tuan and Hui, Siu Cheung and Su, Jian},
  booktitle={Proc. 56th Annu. Meeting Assoc. Comput. Linguist.},
  pages={1010--1020},
  year={2018}
}

@inproceedings{cai2019multi,
  title={Multi-Modal Sarcasm Detection in {Twitter} with Hierarchical Fusion Model},
  author={Cai, Yitao and Cai, Huiyu and Wan, Xiaojun},
  booktitle={Proc. 57th Annu. Meeting Assoc. Comput. Linguist.},
  pages={2506--2515},
  year={2019}
}

@inproceedings{wu2025incongruity,
  title={Incongruity-aware Cross-modal Interaction Network for Multimodal Sarcasm Detection},
  author={Wu, Yujun and Wang, Chen and Chen, Meixuan and Wang, Tongguan and Sha, Ying},
  booktitle={Proc. IEEE Int. Conf. Multimedia Expo},
  pages={1--6},
  year={2025}
}

@inproceedings{xu2020reasoning,
  title={Reasoning with Multimodal Sarcastic Tweets via Modeling Cross-Modality Contrast and Semantic Association},
  author={Xu, Nan and Zeng, Zhixiong and Mao, Wenji},
  booktitle={Proc. 58th Annu. Meeting Assoc. Comput. Linguist.},
  pages={3777--3786},
  year={2020}
}

@article{liang2021cmgcn,
  title={Multi-modal sarcasm detection with interactive graph convolutional network},
  author={Liang, Bin and Lou, Chenwei and Li, Xiang and Gui, Lin and He, Min and Xu, Ruifeng},
  journal={Knowl.-Based Syst.},
  volume={240},
  pages={108101},
  year={2022}
}

@inproceedings{zhu2024tfcd,
  title={{TFCD}: Towards multi-modal sarcasm detection via training-free counterfactual debiasing},
  author={Zhu, Zhihong and Zhuang, Xianwei and Zhang, Yunyan and Xu, Derong and Hu, Guimin and Wu, Xian and Zheng, Yefeng},
  booktitle={Proc. 33rd Int. Joint Conf. Artif. Intell.},
  pages={6687--6695},
  year={2024}
}

@inproceedings{liang2022multimodal,
  title={Multi-modal sarcasm detection via graph convolutional network and dynamic network},
  author={Hao, Jiaqi and Zhao, Junfeng and Wang, Zhigang},
  booktitle={Proc. 33rd ACM Int. Conf. Inf. Knowl. Manage.},
  pages={789--798},
  year={2024}
}

@inproceedings{hazarika2018icon,
  title={{ICON}: Interactive conversational memory network for multimodal emotion detection},
  author={Hazarika, Devamanyu and Poria, Soujanya and Mihalcea, Rada and Cambria, Erik and Zimmermann, Roger},
  booktitle={Proc. Conf. Emp. Methods Natural Lang. Process.},
  pages={2594--2604},
  year={2018}
}

@inproceedings{ref7,
  title={Tensor fusion network for multimodal sentiment analysis},
  author={Zadeh, Amir and Chen, Minghai and Poria, Soujanya and Cambria, Erik and Morency, Louis-Philippe},
  booktitle={Proc. 2017 Conf. Emp. Methods Natural Lang. Process.},
  pages={1103--1114},
  year={2017}
}

@article{eke2021multi,
  title={Multi-feature fusion framework for sarcasm identification on {Twitter} data: A machine learning based approach},
  author={Eke, Christopher Ifeanyi and Norman, Azah Anir and Shuib, Liyana},
  journal={PLOS ONE},
  volume={16},
  number={6},
  pages={e0252918},
  year={2021}
}

@article{yue2023knowlenet,
  title={{KnowleNet}: Knowledge fusion network for multimodal sarcasm detection},
  author={Yue, Tan and Mao, Rui and Wang, Heng and Hu, Zonghai and Cambria, Erik},
  journal={Inf. Fusion},
  volume={100},
  pages={101921},
  year={2023}
}

@inproceedings{ouyang2022cosmo,
  title={Cosmo: contrastive fusion learning with small data for multimodal human activity recognition},
  author={Ouyang, Xiaomin and Shuai, Xian and Zhou, Jiayu and Shi, Ivy Wang and Xie, Zhiyuan and Xing, Guoliang and Huang, Jianwei},
  booktitle={Proc. 28th Annu. Int. Conf. Mob. Comput. Netw.},
  pages={324--337},
  year={2022}
}

@inproceedings{lou2021affective,
  title={Affective dependency graph for sarcasm detection},
  author={Lou, Chenwei and Liang, Bin and Gui, Lin and He, Yulan and Dang, Yixue and Xu, Ruifeng},
  booktitle={Proc. 44th Int. ACM SIGIR Conf. Res. Develop. Inf. Retr.},
  pages={1844--1849},
  year={2021}
}

@inproceedings{singh2024enhancing,
  title={Enhancing Social Media Sarcasm Detection Using Chicken Swarm Optimization and Graph Neural Networks},
  author={Singh, Kamred Udham and Singh, Niharika and Chaudhary, Varun and Paliwal, Deepika and Singh, Teekam and Kumar Dewangan, Amit},
  booktitle={Proc. 2024 IEEE Int. Conf. Contemp. Comput. Commun.},
  volume={1},
  pages={1--6},
  year={2024}
}

@inproceedings{yao2025sarcasm,
  title={Is sarcasm detection a step-by-step reasoning process in large language models?},
  author={Yao, Ben and Zhang, Yazhou and Li, Qiuchi and Qin, Jing},
  booktitle={Proc. 39th AAAI Conf. Artif. Intell.},
  volume={39},
  number={24},
  pages={25651--25659},
  year={2025}
}

@article{zhang2025sarcasmbench,
  title={{Sarcasmbench}: Towards evaluating large language models on sarcasm understanding},
  author={Zhang, Yazhou and Zou, Chunwang and Lian, Zheng and Tiwari, Prayag and Qin, Jing},
  journal={IEEE Trans. Affect. Comput.},
  year={2025},
volume={16},
  number={4},
  pages={2560--2578}
}

@inproceedings{devlin2019bert,
  title={{BERT}:{Pre-training} of deep bidirectional transformers for language understanding},
  author={Devlin, Jacob and Chang, Ming-Wei and Lee, Kenton and Toutanova, Kristina},
  booktitle={Proc. Conf. North Amer.Chapter Assoc. Comput. Linguist. Hum. Lang. Technol},
  pages={4171--4186},
  year={2019}
}

@inproceedings{ref_mustard_plus,
  title={A multimodal corpus for emotion recognition in sarcasm},
  author={Ray, Anupama and Mishra, Shubham and Nunna, Apoorva and Bhattacharyya, Pushpak},
  booktitle={Proc. 13th Lang. Resour. Eval. Conf.},
  pages={6992--7003},
  year={2022}
}

@inproceedings{ref_film,
  title={Filming multimodal sarcasm detection with attention},
  author={Gupta, Sundesh and Shah, Aditya and Shah, Miten and Syiemlieh, Laribok and Maurya, Chandresh},
  booktitle={Proc. Int. Conf. Neural Inf. Process.},
  pages={178--186},
  year={2021}
}

@inproceedings{ref38,
  title={Integrating multimodal information in large pretrained transformers},
  author={Rahman, Wasifur and Hasan, Md Kamrul and Lee, Sangwu and Zadeh, AmirAli Bagher and Mao, Chengfeng and Morency, Louis-Philippe and Hoque, Ehsan},
  booktitle={Proc. 58th Annu. Meeting Assoc. Comput. Linguist.},
  pages={2359--2369},
  year={2020}
}

@inproceedings{liu2021does,
  title={What does your smile mean? {Jointly} detecting multi-modal sarcasm and sentiment using quantum probability},
  author={Liu, Yaochen and Zhang, Yazhou and Li, Qiuchi and Wang, Benyou and Song, Dawei},
  booktitle={Proc. Findings Assoc. Comput. Linguistics: EMNLP 2021},
  pages={871--880},
  year={2021}
}

@inproceedings{tomar2023your,
  title={Your tone speaks louder than your face! {Modality} Order Infused Multi-modal Sarcasm Detection},
  author={Tomar, Mohit and Tiwari, Abhisek and Saha, Tulika and Saha, Sriparna},
  booktitle={Proc. 31st ACM Int. Conf. Multimedia},
  pages={3926--3933},
  year={2023}
}

@article{li2024attention,
  title={An attention-based, context-aware multimodal fusion method for sarcasm detection using inter-modality inconsistency},
  author={Li, Yangyang and Li, Yuelin and Zhang, Shihuai and Liu, Guangyuan and Chen, Yanqiao and Shang, Ronghua and Jiao, Licheng},
  journal={Knowl.- Based Syst.},
  volume={287},
  pages={111457},
  year={2024}
}

@article{pandey2025vyang,
  title={{VyAnG-Net}: A novel multi-modal sarcasm recognition model by uncovering visual, acoustic and glossary features},
  author={Pandey, Ananya and Vishwakarma, Dinesh Kumar},
  journal={Intell. Data Anal.},
  pages={1478--1500},
  year={2025}
}

@article{liang2024fusion,
  title={{Fusion and discrimination}: A multimodal graph contrastive learning framework for multimodal sarcasm detection},
  author={Liang, Bin and Gui, Lin and He, Yulan and Cambria, Erik and Xu, Ruifeng},
  journal={IEEE Trans. Affect. Comput.},
  volume={15},
  number={4},
  pages={1874--1888},
  year={2024}
}

@article{yue2026interarm,
  title={{InterARM}: Interpretable Affective Reasoning Model for Multimodal Sarcasm Detection},
  author={Yue, Tan and Mao, Rui and Shi, Xuzhao and Cambria, Erik},
  journal={IEEE Trans. Affect. Comput.},
  year={2026},
  volume={},
  number={},
  pages={1--12}
}

@ARTICLE{ESAM,
  author={Yuan, Shaozu and Wei, Yiwei and Zhou, Hengyang and Xu, Qinfu and Chen, Meng and He, Xiaodong},
  journal={IEEE Trans. Multimedia}, 
  title={Enhancing Semantic Awareness by Sentimental Constraint With Automatic Outlier Masking for Multimodal Sarcasm Detection}, 
  year={2025},
  volume={27},
  number={},
  pages={5376--5386}
  }

\vfill

\end{document}